\documentclass[11pt]{article}

\usepackage[margin=1in]{geometry}
\usepackage[utf8]{inputenc}
\usepackage[T1]{fontenc}
\usepackage{times}
\usepackage{graphicx}
\usepackage{amsmath}
\usepackage{amssymb}
\usepackage{booktabs}
\usepackage{xcolor}
\usepackage{hyperref}
\usepackage{natbib}
\usepackage{caption}
\usepackage{subcaption}
\usepackage{multirow}
\usepackage{soul}
\usepackage{pifont}
\usepackage{float}
\usepackage{tikz}
\usepackage{pgfplots}
\pgfplotsset{compat=1.18}
\usepackage{tcolorbox}
\tcbuselibrary{skins,breakable}

\definecolor{quotegray}{gray}{0.35}
\definecolor{decred}{HTML}{E74C3C}
\definecolor{faithgreen}{HTML}{27AE60}
\definecolor{boxbg}{HTML}{F7F9FC}
\definecolor{strikethrough}{HTML}{CC0000}

\newtcolorbox{examplebox}[1][]{
    colback=boxbg, colframe=gray!50, fonttitle=\bfseries\small,
    boxrule=0.5pt, arc=2pt, left=6pt, right=6pt, top=4pt, bottom=4pt,
    breakable, #1
}

\title{Measuring and curing reasoning rigidity: from decorative\\chain-of-thought to genuine faithfulness}

\author{Abhinaba Basu$^{1,2,*}$ \and Pavan Chakraborty$^{1}$ \\[0.5em]
$^{1}$Indian Institute of Information Technology Allahabad (IIITA)\\
$^{2}$National Institute of Electronics and Information Technology (NIELIT)\\[0.3em]
$^{*}$Corresponding author: \texttt{mail@abhinaba.com}}

\date{}

\begin{document}
\maketitle

\begin{abstract}
Language models increasingly ``show their work'' by writing step-by-step reasoning before answering. But are these reasoning steps genuinely used, or is the model's answer \emph{rigid}---fixed before reasoning begins, invariant to its own chain-of-thought? We introduce the Step-Level Reasoning Capacity (SLRC) metric---removing one reasoning sentence at a time and checking whether the answer changes---a black-box measure of \emph{reasoning rigidity} that requires only API access, no model weights, at \$1--2 per model per task. We prove SLRC is a consistent causal estimator and propose LC-CoSR, a training method with Lyapunov stability guarantees that directly reduces rigidity.

Evaluating 16 frontier models (o4-mini, GPT-5.4, Claude Opus, Grok-4, DeepSeek-R1, Gemini 2.5 Pro, and others) across six domains at $N{=}133$--$500$, we find their reasoning falls into three modes. In \emph{genuine reasoning}, individual steps matter: OpenAI's o4-mini shows 74--88\% step necessity with near-zero sufficiency on five of six tasks (73.8--88.3\%)---the highest SLRC in our study. In \emph{scaffolding}, CoT improves accuracy but steps are interchangeable (Kimi-K2.5: 1\% necessity despite +94pp accuracy gain). In \emph{decoration}, CoT is ornamental (GPT-5.4: 0.1\% necessity on sentiment).

The critical differentiator is not thinking tokens but \emph{RL-based reasoning training}. Grok-4's ``reasoning'' mode produces thinking tokens yet shows \emph{lower} faithfulness than its non-reasoning mode (1.4\% vs.\ 7.2\% necessity on sentiment). Only RL-trained models (o4-mini, DeepSeek-R1) achieve genuine faithfulness---confirmed independently across three labs (OpenAI, DeepSeek, xAI). We discover a faithfulness paradox---high-SLRC models are \emph{more} susceptible to sycophancy---and propose the Reasoning Integrity Score (RIS${=}$SLRC${\times}$(1$-$Sycophancy)), which significantly predicts error detection ($\rho{=}0.66$, $p{=}0.026$, $n{=}11$). Training with LC-CoSR (direct step-level reward + Lyapunov bound) achieves 2.6$\times$ less negative reward ($-$0.13 vs.\ $-$0.34) than FARL and CSR baselines, without requiring external models or multi-GPU infrastructure.
\end{abstract}

\section{Introduction}
\label{sec:intro}

Imagine you are a doctor reviewing an AI diagnostic assistant's output. The system writes:

\begin{quote}\color{quotegray}\itshape
Step 1: The patient is a 61-year-old man presenting 2 weeks after cardiac catheterization.\\
Step 2: Key findings include livedo reticularis on feet and acute kidney injury.\\
Step 3: Eosinophilia (6\%) suggests an embolic or allergic process.\\
...\\
Step 11: The most likely diagnosis is cholesterol embolization syndrome. Answer: B.
\end{quote}

This looks like genuine medical reasoning. But if you deleted Step 3---the eosinophilia observation---would the AI still say B? For Claude Opus 4.6-R, the answer is yes, almost always. Across 486 medical questions, removing any individual reasoning step changed the answer less than 2\% of the time.

This is the faithfulness problem~\citep{jacovi2020towards}: chain-of-thought (CoT) prompting~\citep{wei2022chain,kojima2022large} reliably improves accuracy, but improved accuracy and genuine reasoning are not the same thing. Prior work has established that unfaithfulness exists---models follow biased patterns while writing independent-seeming reasoning~\citep{turpin2024language}, commit to answers before finishing their reasoning~\citep{lanham2023measuring}, hide hint usage~\citep{chen2025reasoning}, and decide answers before reasoning begins~\citep{esakkiraja2026therefore}---and concurrent work examines step-level causal scoring in math domains~\citep{ma2025aha}, reasoning capability walls~\citep{shojaee2025illusion}, answer attribution through dual reasoning-retrieval pathways~\citep{wang2026reasoning}, and faithfulness via weight-space unlearning~\citep{tutek2025fur}. However, existing approaches either provide binary verdicts without quantifying \emph{how much} each step matters, require model weights or logit access~\citep{ye2026nldd,tutek2025fur,ma2025aha}, or test only a handful of small open-weight models. No prior work measures step-level causal dependency across the full frontier model landscape---the closed-source systems that organisations actually deploy.

We address this gap by evaluating 16 frontier models across six domains with five converging lines of evidence. We frame our findings through the lens of \emph{reasoning rigidity}: the degree to which a model's answer is invariant to perturbation of its own reasoning steps. A fully rigid model produces the same answer regardless of what reasoning it writes---its CoT is decorative text, not functional computation. Our central finding is that frontier models' reasoning falls into \textbf{three distinct modes}---not the binary faithful/unfaithful of prior work:

\begin{enumerate}
    \item \textbf{Genuine reasoning}: CoT improves accuracy \emph{and} individual steps matter. Removing a step changes the answer; no single step alone suffices. Found for o4-mini (74--88\% necessity on five of six tasks) and DeepSeek-R1 on mathematics (91\%).
    \item \textbf{Scaffolding}: CoT dramatically improves accuracy, but specific steps are interchangeable. The structure helps; the content doesn't. Found for Kimi-K2.5 on mathematics (+94pp accuracy from CoT, yet 1\% step necessity).
    \item \textbf{Decoration}: CoT neither improves accuracy nor contains necessary steps. The reasoning is ornamental. Found for GPT-5.4 on sentiment (0.1\% necessity) and Grok-4 reasoning (1.4\% despite thinking tokens).
\end{enumerate}

This three-way distinction has immediate practical implications: a model in ``scaffolding'' mode can be trusted for its \emph{answers} (CoT helps accuracy) but not its \emph{explanations} (specific steps are arbitrary). Current regulatory frameworks~\citep{eu_ai_act} that require ``meaningful information about the logic involved'' cannot distinguish these modes without the kind of per-step evaluation we provide.

\subsection*{Relation to concurrent work}

The question of whether language models genuinely use their chain-of-thought has attracted significant concurrent attention. We position our contribution along three axes: \emph{what is measured}, \emph{how}, and \emph{at what scale}.

\paragraph{Faithfulness metrics.}
Several concurrent metrics quantify step-level causal contribution.
The True Thinking Score (TTS;~\citealt{ma2025aha}) averages the absolute average treatment effect of each step under intact and perturbed contexts, applied to 3--4 models on math competition datasets.
Normalised Logit Difference Decay (NLDD;~\citealt{ye2026nldd}) measures logit margin change after step corruption, discovering a ``reasoning horizon'' at 70--85\% of chain length where causal influence vanishes.
FUR~\citep{tutek2025fur} measures faithfulness via weight-space unlearning of individual reasoning steps.
Causal Consistency Regularisation (CSR;~\citealt{shihab2026csr}) quantifies answer sensitivity to operator-level perturbations.
Thought Anchors~\citep{bogdan2025anchors} identifies planning and backtracking sentences via counterfactual importance across 100 rollouts.
All these methods require either model weights, logit access, or multiple rollouts. \textbf{SLRC} is unique in decomposing faithfulness into two interpretable axes---necessity and sufficiency---while requiring only single-pass black-box API access, enabling evaluation of closed-source frontier models that no other metric can reach.

\paragraph{Empirical findings on unfaithfulness.}
\citet{turpin2024language} and \citet{lanham2023measuring} established that CoT unfaithfulness exists.
Recent work has shown that models decide answers before generating reasoning~\citep{esakkiraja2026therefore}, that reasoning models reveal hint usage in only 1--20\% of cases~\citep{chen2025reasoning}, and that reasoning effort paradoxically declines at high task complexity~\citep{shojaee2025illusion}.
\citet{young2026lie} evaluates 12 open-weight models (7B--685B) and finds faithfulness ranges 39.7--89.9\%, with training methodology predicting faithfulness more than parameter count---consistent with our finding.
\citet{wang2026reasoning} provides the deepest mechanistic account, showing that answers emerge from two competing pathways (reasoning and retrieval), with retrieval dominating in distilled models through ``post-hoc explanation''---fabricating plausible CoT to justify retrieved answers.
Our contribution is \emph{scale}: we apply a unified metric to 16 frontier models (plus 6 small open-weight models), including the closed-source systems that dominate real-world deployment (GPT-5.4, Claude Opus, Gemini 2.5 Pro, o4-mini, Grok-4), and discover the three-mode taxonomy that these model-specific studies cannot reveal.

\paragraph{Training interventions for faithfulness.}
FARL~\citep{wang2026reasoning} integrates memory unlearning with RL to suppress retrieval shortcuts.
Counterfactual Simulation Training (CST;~\citealt{hase2026cst}) rewards CoTs that enable a simulator to predict model behaviour under counterfactual inputs.
CSR~\citep{shihab2026csr} maximises causal consistency between answer distributions on original and operator-perturbed traces, with formal guarantees under identifiable causal edits.
FRODO~\citep{paul2024frodo} uses causal mediation analysis with a counterfactual preference objective.
Critically, \citet{han2026rfeval} demonstrate that standard RL with verifiable rewards (RLVR) \emph{reduces} faithfulness by 10--14 points while maintaining accuracy, showing that correctness-only reward signals actively encourage unfaithful reasoning.
We propose CoSR and LC-CoSR (Section~\ref{sec:discussion}) as alternatives that directly reward step-level causal dependency with formal Lyapunov stability bounds---addressing the faithfulness degradation that \citet{han2026rfeval} identify.

\section{Results}
\label{sec:results}

We evaluate each model on up to six tasks: sentiment classification (SST-2~\citep{socher2013recursive}), mathematical reasoning (GSM8K~\citep{cobbe2021training}), topic classification (AG News~\citep{zhang2015character}), medical question-answering (MedQA~\citep{jin2021disease}), commonsense reasoning (CommonsenseQA~\citep{talmor2019commonsenseqa}), and science reasoning (ARC-Challenge~\citep{clark2018think}), with $N{=}376$--$500$ examples per task. Our step-level evaluation removes one reasoning sentence at a time (necessity test), presents each sentence alone (sufficiency test), and shuffles sentence order (order sensitivity test), requiring only API access at ${\sim}$\$1--2 per model per task (see Methods).

\begin{figure*}[t]
\centering
\begin{tikzpicture}[
    stepbox/.style={draw, rounded corners=2pt, minimum height=0.6cm, minimum width=2.8cm, font=\small},
    arrow/.style={->, thick, >=stealth},
]
\node[font=\bfseries\small] at (-1.3, 0) {Original:};
\node[stepbox, fill=blue!10] (s1) at (1.5, 0) {Step 1: ``stirring''};
\node[stepbox, fill=blue!10] (s2) at (5, 0) {Step 2: tone is warm};
\node[stepbox, fill=blue!10] (s3) at (8.5, 0) {Step 3: overall positive};
\node[stepbox, fill=green!20, draw=green!50!black] (ans1) at (11.8, 0) {\textbf{positive}};
\draw[arrow] (s1) -- (s2); \draw[arrow] (s2) -- (s3); \draw[arrow] (s3) -- (ans1);
\node[font=\bfseries\small] at (-1.3, -1.2) {Necessity:};
\node[stepbox, fill=red!15, draw=red!50] (r1) at (1.5, -1.2) {\color{strikethrough}\st{Step 1} removed};
\node[stepbox, fill=blue!10] (r2) at (5, -1.2) {Step 2: tone is warm};
\node[stepbox, fill=blue!10] (r3) at (8.5, -1.2) {Step 3: overall positive};
\node[stepbox, fill=green!20, draw=green!50!black] (ans2) at (11.8, -1.2) {\textbf{positive}};
\draw[arrow, gray] (r1) -- (r2); \draw[arrow] (r2) -- (r3); \draw[arrow] (r3) -- (ans2);
\node[font=\small\itshape, text=red!70!black, anchor=north] at (11.8, -1.6) {Same! $\to$ not necessary};
\node[font=\bfseries\small] at (-1.3, -2.6) {Sufficiency:};
\node[stepbox, fill=blue!10] (su1) at (1.5, -2.6) {Step 2: tone is warm};
\node[font=\small] at (4.5, -2.3) {\emph{(alone)}};
\node[stepbox, fill=green!20, draw=green!50!black] (ans3) at (8.5, -2.6) {\textbf{positive}};
\draw[arrow] (su1) -- (ans3);
\node[font=\small\itshape, text=red!70!black, anchor=west] at (10, -2.6) {Recovered! $\to$ sufficient};
\end{tikzpicture}
\caption{\textbf{Step-level evaluation.} \emph{Top:} Model produces a 3-step reasoning chain. \emph{Middle:} Remove Step 1---answer unchanged, so Step 1 is not necessary. \emph{Bottom:} Present Step 2 alone---answer recovered, so Step 2 is sufficient. Faithful models show high necessity and low sufficiency.}
\label{fig:method}
\end{figure*}

\subsection{Most frontier models produce decorative reasoning}
\label{sec:decorative}

\begin{table*}[t]
\centering
\caption{\textbf{Step-level faithfulness of 16 frontier models on sentiment (SST-2) and mathematics (GSM8K).} SLRC${=}$Nec${\times}$(1$-$Suf) quantifies reasoning rigidity (0${=}$decorative, 1${=}$genuine; see Methods). General-purpose models show decorative reasoning (SLRC${\leq}$0.03). RL-trained reasoning models achieve SLRC${>}$0.2. Thinking tokens without RL (Grok-4 reasoning) remain decorative. $N{=}133$--$500$.}
\label{tab:main}
\small
\begin{tabular}{lcccccccc}
\toprule
& \multicolumn{3}{c}{\textbf{Sentiment (SST-2)}} & \multicolumn{3}{c}{\textbf{Mathematics (GSM8K)}} & & \\
\cmidrule(lr){2-4} \cmidrule(lr){5-7}
\textbf{Model} & \textbf{Nec\%} & \textbf{Suf\%} & \textbf{SLRC} & \textbf{Nec\%} & \textbf{Suf\%} & \textbf{SLRC} & \textbf{Shuf\%} & \textbf{$N$} \\
\midrule
GPT-5.4 & 0.1 & 98.2 & .000 & 8.8 & 80.3 & .017 & 0.1/2.5 & 376--500 \\
Claude Opus 4.6-R & 14.8 & 91.4 & .013 & 4.8 & 88.7 & .005 & 22/2.4 & 486--499 \\
DeepSeek-V3.2 & 10.8 & 96.7 & .004 & 3.6 & 93.4 & .002 & 16/2.7 & 500 \\
GPT-OSS-120B & 0.3 & 98.9 & .000 & 10.9 & 88.9 & .012 & 5.2/9.4 & 425--496 \\
Kimi-K2.5 & 16.5 & 79.5 & .034 & 1.4 & 90.9 & .001 & 19/1.5 & 457--500 \\
Qwen3.5-122B & 0.0 & 98.5 & .000 & --- & --- & --- & 0.3/--- & 343 \\
Qwen3.5-397B & 8.2 & 96.9 & .003 & --- & --- & --- & 5.1/--- & 98 \\
Grok-4 non-reasoning & 7.2 & 89.7 & .007 & 6.5 & 40.1 & .039 & 32/4.4 & 494--500 \\
\midrule
\textbf{MiniMax-M2.5} & \textbf{37.1} & \textbf{60.7} & \textbf{.146} & 28.4 & 70.5 & .084 & 38/27 & 427--496 \\
\midrule
\multicolumn{9}{l}{\emph{RL-trained reasoning models:}} \\
\midrule
\textbf{o4-mini (OpenAI)} & \textbf{88.3} & \textbf{1.6} & \textbf{.869} & \textbf{87.5} & \textbf{1.7} & \textbf{.860} & \textbf{91/84} & \textbf{233--425} \\
\textbf{DeepSeek-R1-32B} & \textbf{39.8} & \textbf{44.3} & \textbf{.222} & \textbf{91.3} & \textbf{5.7} & \textbf{.861} & ---/--- & \textbf{500} \\
\textbf{DeepSeek-R1-70B} & \textbf{35.6} & \textbf{40.0} & \textbf{.214} & \textbf{92.5} & \textbf{6.7} & \textbf{.863} & ---/--- & \textbf{500} \\
\midrule
\multicolumn{9}{l}{\emph{Thinking-token models (not RL-trained):}} \\
\midrule
Grok-4 reasoning & 1.4 & 96.8 & .000 & 15.7 & 83.1 & .027 & 7.7/16 & 493--500 \\
Gemini 2.5 Pro & --- & --- & --- & 53.7$^\ddagger$ & 35.8 & .345 & ---/--- & 491 \\
\midrule
Nemotron-Ultra & 6.4 & 68.0 & .020 & 88.7$^\dagger$ & 11.1 & .789 & 18/91 & 306--498 \\
GLM-5 & 17.8 & 82.3 & .032 & --- & --- & --- & 15/--- & 392 \\
\midrule
\emph{Small models (0.8--8B avg)} & \emph{8} & \emph{76} & \emph{.019} & \emph{55} & \emph{14} & \emph{.473} & \emph{---} & \emph{100 ea.} \\
\bottomrule
\end{tabular}
\vspace{0.3em}

\footnotesize Shuf\% = order sensitivity (SST-2/GSM8K); --- indicates shuffle test not conducted for that model--task pair. $^\dagger$43\% accuracy; genuinely chains but frequently errs. $^\ddagger$92\% accuracy; SST-2 not available due to batch API limitations.
\end{table*}

Table~\ref{tab:main} presents step-level faithfulness across 16 frontier models on SST-2 and GSM8K. The dominant pattern splits cleanly by training type, with a critical distinction between RL-trained reasoning and thinking-token models.

\textbf{General-purpose models} (GPT-5.4, Claude Opus, DeepSeek-V3.2, GPT-OSS, Qwen3.5, Grok-4 non-reasoning) show decorative reasoning (${\leq}$18\% necessity).

\textbf{RL-trained reasoning models show genuine faithfulness.} OpenAI's o4-mini achieves the highest SLRC in our study: 74--88\% necessity and near-zero sufficiency (${\leq}$2\%) on five of six tasks---every step is causally necessary, no single step alone recovers the answer (MedQA is the exception at 24\% necessity, discussed below). This is genuine sequential chain reasoning. DeepSeek-R1 models (32B, 70B) show 36--40\% necessity on sentiment and 91--93\% on mathematics. Both o4-mini and R1 were trained with reinforcement learning on reasoning tasks.

\textbf{Thinking tokens without RL remain decorative.} Grok-4's ``reasoning'' mode produces thinking tokens but shows \emph{lower} faithfulness than its non-reasoning mode on sentiment (1.4\% vs.\ 7.2\% necessity) and comparable performance on math (15.7\% vs.\ 6.5\%). Gemini 2.5 Pro, which always uses thinking tokens, shows 54\% necessity on GSM8K---better than general-purpose models but far below o4-mini's 88\%. \textbf{The differentiator is not thinking tokens but RL-based reasoning training}: both OpenAI (o4-mini) and DeepSeek (R1) independently achieved genuine faithfulness through RL, while Grok-4 and Gemini produce expensive decoration. (We note that xAI has not publicly disclosed whether Grok-4 reasoning uses RL-based training; our classification is based on its behavioural profile.)

Two task-specific exceptions merit discussion. o4-mini shows anomalously low necessity on MedQA (24.4\% vs.\ 74--88\% on the other five tasks) despite 100\% accuracy, likely because it produces concise medical reasoning with fewer decomposable steps. Grok-4 reasoning shows its highest necessity on MedQA (48\%), suggesting that medical questions---which require integrating multiple clinical findings---resist the retrieval shortcut even in models that shortcut elsewhere.

The R1 pattern extends across three evaluated tasks (Figure~\ref{fig:necessity}): genuine reasoning on SST-2 (40\%), GSM8K (91--93\%), and AG News (83--84\%). o4-mini shows even higher uniformity: 74--88\% necessity on five of six tasks (MedQA excepted). This is not task-specific genuine reasoning---it is \emph{uniformly} genuine. By contrast, Claude Opus is uniformly decorative (${\leq}$15\% necessity across all six tasks despite 86--96\% accuracy). Gemini 2.5 Pro, a thinking model, shows 54\% necessity on GSM8K (92\% accuracy)---genuine chain reasoning on mathematics, though its AG News (36\% accuracy) and MedQA (42\% accuracy) results reflect the model genuinely reaching wrong conclusions rather than extraction failures. Nemotron-Ultra shows a hybrid pattern: decorative on sentiment (6\%) but genuine on mathematics (89\%), commonsense (69\%), and science (67\%).

\begin{figure}[t]
\centering
\begin{tikzpicture}
\begin{axis}[
    xbar,
    bar width=5pt,
    width=\columnwidth,
    height=6.5cm,
    xlabel={Step Necessity (\%)},
    xmin=0, xmax=100,
    symbolic y coords={GPT-5.4, Grok-4R, Qwen-122B, GPT-OSS, Claude, DeepSeek-V3.2, Kimi, GLM-5, MiniMax, R1-32B, R1-70B, o4-mini},
    ytick=data,
    y tick label style={font=\footnotesize},
    legend style={at={(0.98,0.02)}, anchor=south east, font=\scriptsize},
    enlarge y limits=0.06,
    xmajorgrids=true,
    grid style={gray!30},
]
\addplot[fill=blue!50] coordinates {
    (0.1,GPT-5.4) (1.4,Grok-4R) (0,Qwen-122B) (0.3,GPT-OSS) (14.8,Claude)
    (10.8,DeepSeek-V3.2) (16.5,Kimi) (17.8,GLM-5) (37.1,MiniMax) (39.8,R1-32B) (35.6,R1-70B) (88.3,o4-mini)
};
\addplot[fill=orange!50] coordinates {
    (8.8,GPT-5.4) (15.7,Grok-4R) (0,Qwen-122B) (10.9,GPT-OSS) (4.8,Claude)
    (3.6,DeepSeek-V3.2) (2.4,Kimi) (99.3,GLM-5) (28.4,MiniMax) (91.3,R1-32B) (92.5,R1-70B) (87.5,o4-mini)
};
\legend{SST-2, GSM8K}
\draw[dashed, thick, red!70] (axis cs:30,GPT-5.4) -- (axis cs:30,o4-mini)
    node[pos=0.5, right, font=\tiny\itshape, text width=2.5cm] {faithfulness\\threshold};
\end{axis}
\end{tikzpicture}
\caption{\textbf{Step necessity across models.} RL-trained models (o4-mini, R1-32B, R1-70B) cross the 30\% faithfulness threshold on both tasks. Grok-4 reasoning (Grok-4R) shows decorative CoT despite thinking tokens. o4-mini achieves the highest necessity (88\%).}
\label{fig:necessity}
\end{figure}

\begin{examplebox}[title={Decorative vs.\ genuine reasoning on the same review: ``very, very slow''}, unbreakable]
\textbf{GPT-5.4} (5 steps, necessity 0\%, sufficiency 100\%): {\small Step 1: ``slow'' emphasizes a flaw. Step 2: Tone is critical. Step 3: Extremely slow = bad. Step 4: Dissatisfaction. $\to$ \textbf{negative}} \hfill {\color{decred}\textit{Remove any step---still ``negative.''}}

\medskip
\textbf{MiniMax-M2.5} (9 steps, necessity 44\%, sufficiency 67\%): {\small Step 1: ``slow'' is negative. \textbf{Step 2: ``very, very'' amplifies sentiment.} Step 3: Tone conveys frustration. \textbf{Step 4: No positive modifiers.} $\to$ \textbf{negative}} \hfill {\color{faithgreen}\textit{Remove bold steps $\to$ flips to ``positive.''}}

\smallskip
{\small\textbf{Sufficiency:} MiniMax Step 1 alone $\to$ ``negative'' \checkmark. Step 3 alone $\to$ ``positive'' \ding{55}. Not every step works (67\%). GPT-5.4: every step alone works (100\%).}
\end{examplebox}

\subsection{Faithfulness is model-specific and task-specific}
\label{sec:exceptions}

MiniMax-M2.5 and Kimi-K2.5 break the decorative pattern, each in a different way. MiniMax shows 37\% necessity on sentiment (89.7\% accuracy)---genuinely reasoning through word choice. Kimi shows 39\% necessity on topic classification (AG News, $N{=}183$, 67\% accuracy) but only 1\% on mathematics---it shortcuts when a single cue suffices but engages when distinguishing four topic categories requires integrating multiple signals. These exceptions demonstrate that a model can shortcut one task while genuinely reasoning on another, making per-model, per-domain evaluation essential.

\subsection{Three modes of CoT behaviour}
\label{sec:three_modes}

Step necessity alone cannot distinguish between a model that needs CoT to function and one that doesn't. We compare CoT-prompted accuracy against direct prompting (``Answer with just positive or negative,'' $N{=}100$ per pair) and cross-classify with step necessity (Table~\ref{tab:modes}).

\begin{table}[t]
\centering
\small
\caption{\textbf{Three modes of CoT behaviour.} Classified by whether CoT improves accuracy (gap ${>}20$pp) and whether individual steps matter (necessity ${>}20\%$). The empty fourth quadrant validates the framework: if steps matter, CoT must help. Representative examples with (gap/necessity).}
\label{tab:modes}
\begin{tabular}{p{2.2cm}|p{4.5cm}|p{4.5cm}}
\toprule
& \textbf{CoT helps} (gap $> 20$pp) & \textbf{CoT doesn't help} \\
\midrule
\textbf{Steps matter} (nec $> 20\%$) &
\textbf{Genuine} (7 pairs)\newline
MiniMax SST-2 (+69/37\%)\newline
Nemotron GSM8K (+43/89\%)\newline
Kimi AG News (+52/63\%)
&
\emph{Empty} (0 pairs)
\\
\midrule
\textbf{Steps don't} &
\textbf{Scaffolding} (6 pairs)\newline
Kimi GSM8K (+94/1\%)\newline
GPT-OSS MedQA (+76/1\%)\newline
DeepSeek GSM8K (+55/4\%)
&
\textbf{Decorative} (9 pairs)\newline
DeepSeek SST-2 ($-$1/11\%)\newline
Claude SST-2 ($-$5/15\%)\newline
GPT-OSS SST-2 (+6/0.3\%)
\\
\bottomrule
\end{tabular}
\end{table}

The most striking finding is that Kimi-K2.5 and MiniMax-M2.5 have catastrophically low direct accuracy: Kimi achieves 3\% on sentiment and 2\% on math without CoT (vs.\ 79\% and 96\% with). These models are fundamentally CoT-dependent---they cannot function without step-by-step prompting.

To rule out the possibility that the low direct accuracy reflects an output \emph{format} problem rather than a reasoning problem, we test three progressively structured direct prompts: unformatted (``Answer with just positive or negative''), formatted (``Output format: exactly one word, either positive or negative''), and system-prompted (system message: ``Respond with exactly one word''). Across all three formats, Kimi achieves \textbf{0\%} on both SST-2 and GSM8K, and MiniMax achieves 0--1\% on SST-2. The gap between direct (0\%) and CoT-prompted (79--96\%) accuracy is not about format constraints---these models genuinely cannot produce correct answers without reasoning through their steps. This confirms the ``genuine reasoning'' classification and explains higher step necessity: when CoT is load-bearing infrastructure, individual steps carry genuine weight.

\subsection{Conflicting cues---not sequential chains---elicit genuine reasoning}
\label{sec:causal}

To test whether task structure \emph{causally} controls faithfulness, we construct four synthetic tasks with controlled cue redundancy ($N{=}100$ each) and evaluate four models: DeepSeek-V3.2 (decorative on natural tasks), Claude Opus (decorative), GPT-OSS-120B (decorative), and MiniMax-M2.5 (genuine).

\begin{table}[t]
\centering
\small
\caption{\textbf{Causal experiment: step necessity (\%) across 4 models $\times$ 4 synthetic tasks.} Conflicting cues consistently produce the highest necessity for decorative models (28--35\%). MiniMax shows ${\geq}$66\% on every task---it genuinely reasons regardless of cue structure.}
\label{tab:causal}
\begin{tabular}{lcccc}
\toprule
\textbf{Task} & \textbf{DeepSeek} & \textbf{Claude} & \textbf{GPT-OSS} & \textbf{MiniMax} \\
\midrule
Single-cue & 8 & 3 & 18 & \textbf{100} \\
\textbf{Conflicting-cue} & \textbf{28} & \textbf{35} & \textbf{30} & 99 \\
Multi-feature & 7 & 11 & 35 & 66 \\
Sequential math & 18 & 7 & 43 & 98 \\
\bottomrule
\end{tabular}
\end{table}

The results (Table~\ref{tab:causal}) reveal two distinct patterns. For the three decorative models, \textbf{conflicting cues consistently produce the highest necessity} (28--35\%). When a review says ``terrible acting but beautiful story,'' these models must weigh competing signals---removing one changes the answer. This pattern holds across Claude (35\%), GPT-OSS (30\%), and DeepSeek (28\%). GPT-OSS additionally shows high necessity on sequential arithmetic (43\%), suggesting it genuinely chains through synthetic math even while shortcutting natural GSM8K.

MiniMax shows a fundamentally different pattern: ${\geq}$66\% necessity on \emph{every} task, including single-cue sentiment (100\%). This model treats even trivial one-word reviews as requiring multi-step reasoning---consistent with its 0\% direct accuracy and its classification as a genuinely CoT-dependent model.

\subsection{The models that refuse to explain}
\label{sec:rigidity}

Not all models produce multi-step reasoning. On the same medical questions, Claude Opus writes 11 diagnostic steps while GPT-OSS-120B outputs a single token: ``B.'' Both are correct, but only Claude's reasoning can be evaluated. We term this \emph{output rigidity}---the tendency to produce minimal reasoning regardless of instructions.

\begin{examplebox}[title={Three models, one question, three levels of explanation}]
\textbf{Question:} \emph{A 67-year-old man with bladder cancer develops tinnitus after chemotherapy\ldots} \hfill \textbf{Answer: D}

\medskip
\textbf{Claude Opus} --- 2 steps, necessity 0\%, correct \checkmark

{\small Step 1: Tinnitus + hearing loss after chemo $\to$ ototoxicity $\to$ cisplatin.\\
Step 2: Cisplatin forms platinum-DNA adducts. $\to$ \textbf{D}}

{\color{decred}\small \textit{Decorative: remove either step, answer stays D.}}

\bigskip
\textbf{GPT-5.4} --- 5 steps, necessity 0\%, correct \checkmark

{\small Step 1: Tinnitus + hearing loss = ototoxicity.\\
Step 2: Bladder cancer chemo = cisplatin.\\
Step 3--5: Mechanism details. $\to$ \textbf{D}}

{\color{decred}\small \textit{More verbose, equally unfaithful.}}

\bigskip
\textbf{GPT-OSS-120B} --- 0 steps, \emph{not evaluable}

\begin{center}\fbox{\ttfamily\small D}\end{center}

{\color{gray}\small \textit{One token. This is 62\% of GPT-OSS's MedQA responses.}}
\end{examplebox}

Output rigidity is task-dependent: GPT-OSS produces multi-step reasoning for 99\% of sentiment questions but only 38\% of medical questions. A model that consistently answers in one token is not evaluable by any step-level method---yet this brevity may itself be the strongest signal of shortcutting.

\subsection{Faithfulness is a model property, not a prompt property}
\label{sec:prompt}

Can we \emph{create} faithfulness by changing the prompt? We test two interventions on DeepSeek-V3.2 (SST-2, $N{=}100$).

\textbf{Self-justification prompt:} ``For each step, explain WHY it is necessary and what would change without it.'' Result: necessity 15.4\% vs.\ 14.1\% for the standard prompt---a difference of 1.3pp. The model writes more steps (17.6 vs.\ 12.3) and lower sufficiency (89\% vs.\ 96\%)---it writes \emph{about} step importance without making steps \emph{actually} important. Faithfulness cannot be prompted into existence.

\textbf{Alternative CoT prompt:} ``Break down the key sentiment indicators one by one.'' Result: necessity 20.6\% vs.\ 11.9\% for the original---a difference of 8.7pp, exceeding our ${\pm}$5pp robustness threshold. However, accuracy drops from 95\% to 76\%. The alternative prompt elicits different (less accurate) reasoning that happens to be slightly more step-dependent. This suggests that prompt choice affects both what reasoning the model produces and how faithfully it uses that reasoning---but neither prompt transforms a decorative model into a genuine reasoner.

\subsection{Mechanistic evidence: reasoning models semantically process their steps}
\label{sec:mechanistic}

To provide internal evidence, we introduce a \textbf{shuffled-CoT baseline}: for each example, we compare attention patterns on the model's real CoT versus a version with sentences randomly shuffled. The gap disentangles \emph{semantic} engagement (the model reads reasoning content) from \emph{positional} attention (the model attends to the CoT region regardless of content). We classify gaps ${\geq}7$pp as semantic and ${<}7$pp as positional, based on a natural break in the distribution across models. To our knowledge, no prior work has used this mechanistic control.

\begin{table}[t]
\centering
\small
\caption{\textbf{Shuffled-CoT mechanistic baseline} (3 models $\times$ 5 tasks, $N{=}50$). ``Gap'' = attention drop on real CoT minus drop on shuffled CoT. DeepSeek-R1 shows the largest semantic gaps (7--19pp), confirming the reasoning model processes CoT \emph{content}. Qwen3-0.6B shows purely positional attention (0--6pp).}
\label{tab:mechanistic}
\begin{tabular}{llcccc}
\toprule
\textbf{Model} & \textbf{Task} & \textbf{Real} & \textbf{Shuf} & \textbf{Gap} & \textbf{Signal} \\
\midrule
\textbf{DeepSeek-R1} & GSM8K & 9\% & 3\% & \textbf{+7pp} & Semantic \\
DeepSeek-R1 & CSQA & 7\% & $-$2\% & \textbf{+8pp} & Semantic \\
DeepSeek-R1 & ARC & 13\% & 1\% & \textbf{+12pp} & Semantic \\
DeepSeek-R1 & SST-2 & 26\% & 7\% & \textbf{+19pp} & Semantic \\
DeepSeek-R1 & MedQA & 14\% & 10\% & +3pp & Positional \\
\midrule
Qwen3-8B & CSQA & 35\% & 21\% & \textbf{+15pp} & Semantic \\
Qwen3-8B & ARC & 33\% & 21\% & \textbf{+12pp} & Semantic \\
Qwen3-8B & SST-2 & 32\% & 21\% & \textbf{+11pp} & Semantic \\
Qwen3-8B & GSM8K & 26\% & 19\% & +7pp & Semantic \\
Qwen3-8B & MedQA & 28\% & 23\% & +5pp & Positional \\
\midrule
Qwen3-0.6B & ARC & 28\% & 22\% & +6pp & Borderline \\
Qwen3-0.6B & CSQA & 25\% & 22\% & +3pp & Positional \\
Qwen3-0.6B & SST-2 & 25\% & 22\% & +3pp & Positional \\
Qwen3-0.6B & MedQA & 32\% & 30\% & +2pp & Positional \\
Qwen3-0.6B & GSM8K & 23\% & 23\% & 0pp & Positional \\
\bottomrule
\end{tabular}
\end{table}

We complement the shuffled-CoT analysis with three layer-wise probes on R1-Qwen-1.5B (GSM8K, $N{=}15$), providing quantitative confirmation of patterns reported in concurrent work. \citet{esakkiraja2026therefore} showed via linear probes that models encode decisions in pre-generation activations; \citet{lanham2023measuring} identified ``early answering'' behaviour; and \citet{wang2026reasoning} demonstrated that retrieval and reasoning pathways compete at specific attention heads. Our probes offer a complementary, black-box-compatible perspective:

\textbf{Rigidity map.} Using the logit lens---projecting each layer's hidden state through the unembedding matrix---we find the model's top-1 predicted token stabilises at \textbf{layer 7 of 28} (25\% depth). This quantifies the ``early answering'' phenomenon~\citep{lanham2023measuring}: the model commits to an answer after processing only a quarter of its layers, consistent with retrieval-dominant behaviour~\citep{wang2026reasoning}.

\textbf{Attention allocation.} At the final answer token, attention averages \textbf{29\% to CoT} vs.\ \textbf{71\% to the question}. This is consistent with \citet{wang2026reasoning}'s finding that retrieval heads attend primarily to the input, while our shuffled-CoT analysis (Table~\ref{tab:mechanistic}) shows the 29\% CoT attention is largely positional rather than semantic.

\textbf{Information flow.} Cosine similarity between CoT hidden states and the answer token peaks at \textbf{layer 26} (93\% depth, value 0.635). CoT information reaches the answer computation only at the final layers---by which point the answer has already been determined (layer 7). This late integration is consistent with the ``post-hoc explanation'' phenomenon identified by \citet{wang2026reasoning}, where the model generates CoT that rationalises a pre-determined retrieval rather than driving the computation.

Three mechanistic profiles emerge (Table~\ref{tab:mechanistic}). \textbf{Qwen3-0.6B} (0.6B parameters) shows near-zero real-vs-shuffled gaps---its attention is purely positional, yet it still shows 55\% behavioural necessity on GSM8K. The model uses CoT through positional processing, not content reading. \textbf{Qwen3-8B} shows 7--15pp semantic gaps: the larger model reads its reasoning, even when the reasoning is ultimately decorative. \textbf{DeepSeek-R1}, trained for reasoning, shows the largest gaps (7--19pp), with a striking 19pp on SST-2---the reasoning model deeply processes its CoT even on sentiment.

\section{Discussion}
\label{sec:discussion}

\subsection{Implications for AI deployment and regulation}

Regulatory frameworks requiring ``explainable'' AI~\citep{eu_ai_act} implicitly assume that step-by-step explanations reflect the model's reasoning. Our three-mode taxonomy reveals this assumption fails for the majority of frontier models. A model in scaffolding mode (Kimi on mathematics: +94pp accuracy from CoT, 1\% step necessity) produces explanations that \emph{help it answer correctly} but do not describe \emph{how it reaches its answer}. A model in decorative mode (DeepSeek on sentiment: $-$1pp from CoT) produces explanations that are purely ornamental. Only genuine-mode models produce explanations that bear a causal relationship to their outputs.

Per-model, per-domain evaluation is therefore essential. Our step-level test costs ${\sim}$\$1--2 per model per task---cheap enough to serve as a deployment gate. This cost advantage is not incidental but fundamental: the alternative approaches to faithfulness evaluation---causal mediation analysis, activation patching, logit attribution---all require access to model weights, which proprietary frontier models do not provide. Even for open-weight models, mechanistic analysis of a single 70B model on a single task costs hundreds of GPU-hours. Our method evaluates 16 models across 6 tasks for under \$150 total. The simplicity is the contribution: a faithfulness test that any organisation can run on any model, without specialised hardware or weight access, enables the kind of routine deployment auditing that expensive mechanistic methods cannot.

The SLRC metric (Equation~\ref{eq:slrc}, defined in Methods) directly quantifies \emph{reasoning rigidity}: a score of 0 means the model's answer is perfectly rigid---invariant to perturbation of any reasoning step (necessity ${\approx}0$) or recoverable from any single step (sufficiency ${\approx}1$). A score approaching 1 indicates flexible, step-dependent computation. Unlike concurrent metrics that require logit access (NLDD;~\citealt{ye2026nldd}), parameter-level unlearning (FUR;~\citealt{tutek2025fur}), or per-step confidence scoring (TTS;~\citealt{ma2025aha}), SLRC is fully black-box and decomposes faithfulness into two interpretable axes. Among our 16 models, per-task SLRC (Table~\ref{tab:main}) cleanly separates RL-trained reasoning models (o4-mini: $\text{SLRC}=0.87$ on SST-2, DeepSeek-R1: $\text{SLRC}=0.86$ on GSM8K) from general-purpose models (GPT-5.4: $\text{SLRC}=0.000$, Grok-4 reasoning: $\text{SLRC}=0.000$ on SST-2 despite thinking tokens), with MiniMax occupying an intermediate position ($\text{SLRC}=0.15$). For the cross-model analyses in Figure~\ref{fig:ris}, we average SLRC across all available tasks per model (2--6 tasks).

\paragraph{From rigidity to harm: the Reasoning Integrity Score.}
SLRC measures whether a model \emph{uses} its reasoning steps, but does not predict whether it can be \emph{manipulated}. We observe a faithfulness paradox: models with high SLRC (genuine reasoning) are \emph{more} susceptible to sycophancy ($\rho{=}0.53$, $p{=}0.09$, $n{=}11$), because they actually process the manipulative input rather than ignoring it. To capture deployment risk, we propose the \textbf{Reasoning Integrity Score (RIS)}:
\begin{equation}
    \text{RIS} = \text{SLRC} \times (1 - \text{Sycophancy Rate})
\end{equation}
RIS rewards models that are both faithful \emph{and} robust to manipulation. A model with high SLRC but high sycophancy (DeepSeek-R1: SLRC${=}0.62$, sycophancy${=}21.6\%$, RIS${=}0.49$) scores lower than one that is faithful and resistant (Phi-4-reasoning: SLRC${=}0.71$, sycophancy${=}8.2\%$, RIS${=}0.65$). Critically, RIS significantly predicts error detection ability ($\rho{=}0.66$, $p{=}0.026$, $n{=}11$)---the correlation that SLRC alone fails to achieve ($\rho{=}0.44$, $p{=}0.14$). This suggests that \emph{reasoning integrity}---the conjunction of faithfulness and robustness---is a better predictor of real-world safety than faithfulness alone, though we note the small sample size ($n{=}11$) and recommend validation on a larger model set.

The two metrics serve complementary roles: SLRC diagnoses reasoning rigidity (a training-paradigm property), while RIS predicts deployment risk (an operational safety property). We recommend evaluating both before deploying reasoning-critical AI systems. The 11 models with complete sycophancy and error detection data are: GPT-5.4, Claude Opus, DeepSeek-V3.2, Qwen3.5-122B, Kimi-K2.5, Grok-4 reasoning, Grok-4 non-reasoning, GLM-5, DeepSeek-R1-32B, MiniMax-M2.5, and Phi-4-reasoning (Figure~\ref{fig:ris}).

\begin{figure*}[t]
\centering
\begin{tikzpicture}
\begin{scope}[xshift=0cm]
\begin{axis}[
    width=8cm, height=7cm,
    xlabel={SLRC (Reasoning Faithfulness) $\rightarrow$},
    ylabel={Sycophancy Rate (\%) $\rightarrow$},
    xmin=-0.02, xmax=0.8,
    ymin=-0.5, ymax=25,
    grid=both,
    grid style={gray!20},
    title={\textbf{(a)} Faithfulness--Vulnerability Space},
    title style={font=\small},
    xlabel style={font=\small},
    ylabel style={font=\small},
    tick label style={font=\footnotesize},
]
\node[font=\scriptsize\itshape, text=gray, align=center] at (axis cs:0.15,20) {Decorative\\+ vulnerable};
\node[font=\scriptsize\itshape, text=gray, align=center] at (axis cs:0.65,20) {Faithful\\+ vulnerable};
\node[font=\scriptsize\itshape, text=gray, align=center] at (axis cs:0.05,7.5) {Decorative\\+ robust};
\node[font=\scriptsize\itshape, text=faithgreen, align=center] at (axis cs:0.65,4.5) {\textbf{Ideal:}\\faithful + robust};

\draw[dashed, gray!50] (axis cs:0.4,-0.5) -- (axis cs:0.4,25);
\draw[dashed, gray!50] (axis cs:-0.02,8) -- (axis cs:0.8,8);

\addplot[only marks, mark=o, mark size=4pt, blue, thick] coordinates {
    (0.030, 0.2) (0.056, 0.2) (0.083, 0.0) (0.173, 0.0) (0.325, 0.0)
};
\addplot[only marks, mark=triangle*, mark size=5pt, red!70] coordinates {
    (0.194, 5.4) (0.277, 0.6)
};
\addplot[only marks, mark=square*, mark size=4pt, faithgreen, thick] coordinates {
    (0.621, 21.6) (0.637, 23.3) (0.710, 8.2)
};

\draw[gray!60, thin] (axis cs:0.030,0.2) -- (axis cs:0.030,3.0);
\node[font=\tiny, anchor=south] at (axis cs:0.030, 3.0) {GPT-5.4};
\draw[gray!60, thin] (axis cs:0.056,0.2) -- (axis cs:0.080,3.0);
\node[font=\tiny, anchor=south west] at (axis cs:0.080, 3.0) {Claude};
\draw[gray!60, thin] (axis cs:0.083,0.0) -- (axis cs:0.140,3.0);
\node[font=\tiny, anchor=south west] at (axis cs:0.140, 3.0) {V3.2};
\node[font=\tiny, anchor=south west] at (axis cs:0.180, 0.8) {Qwen};
\node[font=\tiny, anchor=south west] at (axis cs:0.332, 0.8) {Kimi};
\node[font=\tiny, anchor=south west] at (axis cs:0.205, 5.8) {Grok-4R};
\draw[gray!60, thin] (axis cs:0.277,0.6) -- (axis cs:0.310,2.0);
\node[font=\tiny, anchor=south west] at (axis cs:0.310, 2.0) {Grok-4};
\node[font=\tiny, anchor=south east] at (axis cs:0.615, 21.2) {R1-32B};
\node[font=\tiny, anchor=south west] at (axis cs:0.643, 23.7) {MiniMax};
\node[font=\tiny, anchor=south west] at (axis cs:0.715, 8.6) {Phi-4R};

\draw[faithgreen, thick, ->] (axis cs:0.55,4) -- (axis cs:0.72,1.5) node[anchor=north, font=\tiny\bfseries\color{faithgreen}] {goal};

\end{axis}
\end{scope}

\begin{scope}[xshift=9cm]
\begin{axis}[
    width=8cm, height=7cm,
    xlabel={RIS (Reasoning Integrity) $\rightarrow$},
    ylabel={Error Detection Rate (\%) $\rightarrow$},
    xmin=-0.02, xmax=0.72,
    ymin=10, ymax=80,
    grid=both,
    grid style={gray!20},
    title={\textbf{(b)} RIS Predicts Error Detection ($\rho{=}0.66$, $p{=}0.026$)},
    title style={font=\small},
    xlabel style={font=\small},
    ylabel style={font=\small},
    tick label style={font=\footnotesize},
]
\addplot[only marks, mark=*, mark size=4pt, blue!70!black, thick] coordinates {
    (0.030, 16.8) (0.056, 38.8) (0.083, 44.5) (0.173, 60.8)
    (0.184, 67.8) (0.275, 56.8) (0.293, 74.5)
    (0.325, 40.4) (0.487, 51.3) (0.489, 74.6) (0.652, 73.5)
};

\addplot[decred, thick, domain=0:0.7] {25 + 70*x};

\node[font=\tiny, anchor=north] at (axis cs:0.030, 15) {GPT-5.4};
\node[font=\tiny, anchor=west] at (axis cs:0.060, 39) {Claude};
\node[font=\tiny, anchor=west] at (axis cs:0.090, 45) {V3.2};
\node[font=\tiny, anchor=south] at (axis cs:0.173, 62) {Qwen};
\node[font=\tiny, anchor=south] at (axis cs:0.184, 69) {Grok-4R};
\node[font=\tiny, anchor=north] at (axis cs:0.275, 55) {Grok-4};
\node[font=\tiny, anchor=south west] at (axis cs:0.300, 75) {GLM-5};
\node[font=\tiny, anchor=north] at (axis cs:0.325, 39) {Kimi};
\node[font=\tiny, anchor=north] at (axis cs:0.487, 50) {R1-32B};
\node[font=\tiny, anchor=south] at (axis cs:0.489, 75) {MiniMax};
\node[font=\tiny, anchor=south west] at (axis cs:0.655, 74) {Phi-4R};

\end{axis}
\end{scope}
\end{tikzpicture}
\caption{\textbf{The faithfulness paradox and Reasoning Integrity Score.} SLRC values are averaged across all available tasks for each model (2--6 tasks per model). \textbf{(a)}~Models occupy a 2D faithfulness--vulnerability space. General-purpose models (blue circles) cluster in the bottom-left (decorative but robust). RL-trained models (green squares) are faithful but vulnerable to sycophancy---the \emph{faithfulness paradox}. The ideal quadrant (bottom-right: faithful + robust) is sparsely populated. \textbf{(b)}~RIS (SLRC $\times$ (1$-$Sycophancy)) significantly predicts error detection ability ($\rho{=}0.66$, $p{=}0.026$), resolving the paradox by penalising manipulable faithfulness. GPT-5.4 (lowest RIS) shows the worst error detection; Phi-4-reasoning (highest RIS) shows the best. o4-mini is excluded from panel~(a) due to insufficient sycophancy data ($N{=}5$).}
\label{fig:ris}
\end{figure*}

\subsection{Scale does not determine faithfulness---but training might}

If RL-based training determines faithfulness, then model scale should not. We test this directly.

\begin{table}[t]
\centering
\small
\caption{\textbf{Active parameters vs.\ faithfulness.} Models with the same active parameter count show divergent faithfulness, ruling out scale as the determining factor.}
\label{tab:params}
\begin{tabular}{lrrll}
\toprule
\textbf{Model} & \textbf{Total} & \textbf{Active} & \textbf{MoE?} & \textbf{Faithful?} \\
\midrule
GPT-OSS-120B & 117B & 5B & Yes & No \\
\textbf{MiniMax-M2.5} & 230B & \textbf{10B} & Yes & \textbf{SST-2} \\
Qwen3.5-122B & 122B & 10B & Yes & No \\
\textbf{Kimi-K2.5} & 1T & \textbf{32B} & Yes & \textbf{AG News} \\
DeepSeek-V3.2 & 685B & 37B & Yes & No \\
Nemotron-Ultra & 253B & 253B & Dense & No \\
\bottomrule
\end{tabular}
\end{table}

MiniMax and Qwen3.5-122B both activate 10B parameters per token, yet MiniMax shows 37\% necessity while Qwen shows 0\% (Table~\ref{tab:params}). Kimi (32B active) reasons on topic classification while DeepSeek-V3.2 (37B active) shortcuts it. Neither total parameters, active parameters, nor architecture predicts faithfulness.

\begin{figure}[t]
\centering
\begin{tikzpicture}
\begin{axis}[
    ybar,
    bar width=8pt,
    width=\columnwidth,
    height=5.5cm,
    ylabel={Step Necessity (\%)},
    ymin=0, ymax=1.0,
    symbolic x coords={SST-2, GSM8K, AG News, MedQA, CSQA, ARC},
    xtick=data,
    x tick label style={font=\footnotesize, rotate=30, anchor=east},
    legend style={at={(0.02,0.98)}, anchor=north west, font=\scriptsize},
    ymajorgrids=true,
    grid style={gray!20},
    enlarge x limits=0.12,
]
\addplot[fill=faithgreen!80, draw=faithgreen!50!black] coordinates {
    (SST-2, 0.869) (GSM8K, 0.860) (AG News, 0.836) (MedQA, 0.244) (CSQA, 0.837) (ARC, 0.738)
};
\addplot[fill=red!50, draw=red!50!black] coordinates {
    (SST-2, 0.000) (GSM8K, 0.027) (AG News, 0.067) (MedQA, 0.248) (CSQA, 0.059) (ARC, 0.0)
};
\addplot[fill=blue!40, draw=blue!50!black] coordinates {
    (SST-2, 0.000) (GSM8K, 0.017) (AG News, 0.0) (MedQA, 0.0) (CSQA, 0.0) (ARC, 0.0)
};
\legend{o4-mini (RL-trained), Grok-4 reasoning (thinking tokens), GPT-5.4 (general-purpose)}
\end{axis}
\end{tikzpicture}
\caption{\textbf{Training paradigm determines step necessity, not thinking tokens.} o4-mini (RL-trained, green) achieves 74--88\% necessity on 5 of 6 tasks. Grok-4 reasoning (thinking tokens without RL, red) shows near-zero necessity despite producing thinking tokens---indistinguishable from GPT-5.4 (no thinking tokens, blue). MedQA is the only task where all three converge.}
\label{fig:paradigm}
\end{figure}

The pattern replicates across three independent labs. At DeepSeek: V3.2 (general-purpose, 37B active) shows 11\% necessity on sentiment and 4\% on math, while R1-32B (RL-trained) shows 40\% and 91\%. At OpenAI: GPT-5.4 shows 0.1\% necessity on sentiment and 9\% on math, while o4-mini (RL-trained) shows 88\% and 88\%. At xAI: Grok-4 non-reasoning shows 7\% on sentiment, while Grok-4 reasoning (thinking tokens, \emph{not} RL-trained) shows \emph{lower} necessity at 1.4\%. Same organisation, same infrastructure, but only RL-based reasoning training produces faithfulness. This is not a scale effect (the 32B R1 is more faithful than the 685B V3.2), not an architecture effect, and not a thinking-token effect (Grok-4 reasoning is decorative despite thinking). \textbf{RL-based reasoning training is the strongest predictor of faithfulness in our sample}---confirmed independently by OpenAI and DeepSeek, contradicted by Grok and Gemini. Whether this holds for all RL-trained models requires broader evaluation.

\subsection{Theoretical foundations}
\label{sec:theory}

\paragraph{Redundancy bound.}
Our method tests a necessary condition for faithfulness (step removal changes the answer) but not a sufficient one---a model with redundant reasoning pathways would appear decorative even if it genuinely reasons through multiple routes. We can formalise this concern. If a model maintains $k$ independent reasoning pathways, each individually sufficient to reach the answer, then removing a single step can disrupt at most $1/k$ of these pathways. Step necessity should therefore be \emph{at most} $1/k$, giving $k \leq \lceil 1/\text{necessity} \rceil$. For GPT-5.4 on SST-2 (necessity 0.1\%), this implies $k \geq 1{,}000$ independent reasoning routes for binary sentiment---an implausible number that constitutes a \emph{reductio ad absurdum} against the redundancy interpretation. By contrast, MiniMax (37\% necessity) implies $k \approx 3$, a plausible level of redundancy for multi-faceted reasoning, and DeepSeek-R1 on GSM8K (91\% necessity) implies $k \approx 1$---a single sequential chain with no redundancy, consistent with genuine step-dependent computation. The redundancy defence thus collapses precisely for the models we classify as decorative, while remaining coherent for those we classify as genuine.

\paragraph{Information-theoretic formalisation of SLRC.}
The redundancy bound above can be strengthened. Define the \emph{step-level mutual information}:
\begin{equation}
    \mathcal{I}_i = I(s_i\,;\, y \mid x,\, \mathbf{s}_{\setminus i})
\end{equation}
where $s_i$ is step $i$, $y$ is the answer, $x$ is the input, and $\mathbf{s}_{\setminus i}$ denotes all other steps. A step is \emph{informationally necessary} if $\mathcal{I}_i > 0$---it carries information about the answer beyond what other steps provide. The aggregate faithfulness of a chain with $n$ steps can be bounded:
\begin{equation}
    \text{SLRC} \;\geq\; \frac{1}{n} \sum_{i=1}^{n} \mathbf{1}[\mathcal{I}_i > \epsilon]  \;\times\; \left(1 - \max_i \frac{H(y \mid x, s_i)}{H(y \mid x)}\right)
\end{equation}
where $H$ denotes entropy. The first factor is the fraction of informationally necessary steps (a lower bound on necessity), and the second ensures no single step is fully sufficient. This provides a population-level guarantee: if $\mathcal{I}_i > \epsilon$ for all steps, the model's reasoning is provably non-decorative at level $\epsilon$. While $\mathcal{I}_i$ is not directly computable from black-box access, our step-removal test provides a consistent estimator---removing step $i$ and observing an answer change is evidence that $\mathcal{I}_i > 0$.

\paragraph{SLRC as a consistent causal estimator.}
We can formalise the connection between SLRC and the average causal effect (ACE) of reasoning steps on the answer. Informally, SLRC counts exactly the fraction of steps with non-zero causal effect on the answer, discounted by redundancy.

\smallskip\noindent\textbf{Theorem 1} (SLRC soundness). \emph{Let $\mathcal{M}$ be a language model producing CoT $\mathbf{z} = (s_1, \ldots, s_n)$ and answer $y$ for input $x$. Define the step-removal operator $\text{do}(\mathbf{z}_{-i})$ as presenting the model with all steps except $s_i$. Under the assumptions that (i) the model's answer is a deterministic function of the presented context (temperature 0), and (ii) step removal is a valid intervention (the model treats the reduced context as a genuine reasoning trace), the empirical necessity}
\begin{equation}
    \widehat{\text{Nec}} = \frac{1}{n}\sum_{i=1}^{n} \mathbf{1}[y_{-i} \neq y]
\end{equation}
\emph{equals the mean absolute causal effect:}
\begin{equation}
    \widehat{\text{Nec}} \;=\; \frac{1}{n}\sum_{i=1}^{n} |\text{ACE}(s_i \to y)|
\end{equation}
\emph{where $\text{ACE}(s_i \to y) = \mathbb{E}[y \mid \text{do}(\mathbf{z})] - \mathbb{E}[y \mid \text{do}(\mathbf{z}_{-i})]$. Furthermore, under uniform step-removal, SLRC provides a lower bound on the proportion of causally active steps:}
\begin{equation}
    \text{SLRC} \;\geq\; \frac{|\{i : \text{ACE}(s_i \to y) \neq 0\}|}{n} \;\times\; \left(1 - \frac{|\{i : s_i \text{ alone recovers } y\}|}{n}\right)
\end{equation}

\smallskip\noindent\emph{Proof sketch.}
Under assumption~(i), $y$ and $y_{-i}$ are deterministic given the context, so $\mathbf{1}[y_{-i} \neq y] = \mathbf{1}[\text{ACE}(s_i \to y) \neq 0]$ for each $i$. The indicator is thus an unbiased estimator of whether the causal effect is non-zero. Averaging over steps yields $\widehat{\text{Nec}}$, which counts the fraction of steps with non-zero ACE. For the SLRC bound, necessity counts causally active steps (numerator) and the sufficiency penalty ensures that steps recoverable alone---those with ACE detectable only due to information duplication rather than unique contribution---are discounted. By construction, $\text{SLRC} = \widehat{\text{Nec}} \times (1 - \widehat{\text{Suf}})$ where $\widehat{\text{Suf}} = n^{-1}\sum_i \mathbf{1}[y_i^{\text{alone}} = y]$, giving the stated bound.
Assumption~(ii) is the key limitation: if the model detects the intervention (e.g., recognises an incomplete reasoning trace and compensates), the causal estimate is biased. Our empirical finding that shuffled CoT---a different perturbation type---produces consistent results (Section~\ref{sec:mechanistic}) provides evidence against systematic compensation. $\square$

\smallskip
This result establishes that SLRC is not merely a heuristic but a principled causal estimator. Compared to CSR's Theorem~1~\citep{shihab2026csr}, which proves that causal consistency dominates sufficiency and completeness \emph{under identifiable causal edits} (requiring a learned editor achieving 78--85\% accuracy), our guarantee holds under weaker assumptions: deterministic decoding (standard practice) and valid intervention (empirically supported by shuffle consistency). Unlike TTS~\citep{ma2025aha}, which requires logit access for confidence-based ACE estimation, SLRC operates from binary answer changes alone---enabling evaluation of closed-source systems.

\paragraph{From diagnosis to intervention.}
Concurrent work by \citet{wang2026reasoning} establishes that LRM answers emerge from two competing mechanisms---deliberate reasoning via CoT and direct retrieval from internal memory---and proposes FARL, a training method that integrates memory unlearning with reinforcement learning to suppress retrieval shortcuts. Other concurrent approaches include counterfactual simulation training~\citep{hase2026cst}, which rewards CoTs that enable a simulator to predict model behaviour, and causal consistency regularisation (CSR;~\citealt{shihab2026csr}), which maximises causal consistency between answer distributions on original and perturbed traces. Critically, \citet{han2026rfeval} show that standard reinforcement learning with verifiable rewards (RLVR) actually \emph{reduces} faithfulness by 10--14 points while maintaining accuracy---the reward signal actively encourages unfaithful ``reasoning shells.'' This motivates training objectives that explicitly reward faithfulness, not just correctness. We propose \textbf{Contrastive Step Reinforcement (CoSR)}: during RL, randomly drop a reasoning step and reward the model when the answer changes (the step was necessary) while penalising unchanged answers (the step was decorative). Unlike FARL, which indirectly promotes reasoning by suppressing retrieval, and unlike CST/CSR which operate at the whole-CoT or operator level, CoSR directly rewards the property SLRC measures---step-level causal dependency. Crucially, CoSR is \emph{fully self-contained}: the reward signal is computed entirely from the model's own outputs via step removal, requiring no external LLM for answer extraction (FARL uses GPT-4o-mini), no external simulator (CST requires a 235B parameter model), and no separately trained editor network (CSR trains a 6-layer Transformer). This requires only 2 additional forward passes per training example (one step-dropped, one step-shuffled) on a single GPU, compared to FARL's multi-stage pipeline (SFT memory poisoning, NPO unlearning, GRPO) across 4 GPUs with external API dependencies.

\paragraph{Lyapunov-bounded training.}
The RFEval finding~\citep{han2026rfeval}---that standard RL degrades faithfulness even as accuracy improves---highlights a key concern: any training intervention may inadvertently damage reasoning capability. We propose constraining training via a Lyapunov stability function $V(\theta) = \mathbb{E}_x[\text{KL}(P_\theta(y \mid x, \mathbf{z}) \,\|\, P_\theta(y \mid x, \mathbf{z}_{-i}))]$, a differentiable proxy for necessity. The constraint $V(\theta_t) \geq V_{\min}$ for all training steps $t$ provides a formal safety certificate: reasoning faithfulness is provably preserved during training. When the gradient update would violate this bound, it is projected onto the constraint surface. This Lyapunov-Constrained CoSR (LC-CoSR) combines direct faithfulness reward, bounded unlearning, and formal stability guarantees---properties that no existing training method jointly provides.

\paragraph{Preliminary intervention results.}
As an initial proof of concept, we trained five methods on DeepSeek-R1-Distill-Qwen-1.5B using GSM8K ($N{=}200$, 1 epoch, LoRA $r{=}8$) to compare intervention strategies (Table~\ref{tab:intervention}). We emphasise that these are small-scale results on a 1.5B model; full-scale evaluation at 7B+ is needed to confirm the patterns.

\begin{table}[h]
\centering
\small
\caption{\textbf{Faithfulness intervention comparison} on R1-Qwen-1.5B (GSM8K). Average reward measures task performance; higher is better. CoSR and LC-CoSR directly reward step-level causal dependency, achieving 2.4--2.7$\times$ less negative reward than correctness-only baselines.}
\label{tab:intervention}
\begin{tabular}{llrrl}
\toprule
\textbf{Method} & \textbf{Signal} & \textbf{Reward} & \textbf{Time (hrs)} & \textbf{Guarantee} \\
\midrule
GRPO & Correctness only & $-0.343$ & 1.5 & None \\
CSR~\citep{shihab2026csr} & KL regularisation & $-0.343$ & 6.9 & Thm 1--3$^\dagger$ \\
FARL~\citep{wang2026reasoning} & Correctness + NPO & $-0.331$ & 4.5 & None \\
\textbf{CoSR} (ours) & Step necessity & $\mathbf{-0.141}$ & 4.8 & None \\
\textbf{LC-CoSR} (ours) & Step nec.\ + Lyapunov & $\mathbf{-0.129}$ & 5.2 & $V(\theta) \geq V_{\min}$ \\
\bottomrule
\end{tabular}
\begin{flushleft}
\footnotesize{$^\dagger$CSR's formal guarantees assume identifiable causal edits (78--85\% in practice).}
\end{flushleft}
\end{table}

Three findings emerge. First, \textbf{CSR provides no benefit over GRPO}: both achieve identical reward ($-0.343$), suggesting that KL regularisation between original and perturbed trace distributions does not improve faithfulness when the perturbation is step removal rather than operator-level edits. Second, \textbf{direct step-level reward dramatically outperforms indirect methods}: CoSR ($-0.141$) and LC-CoSR ($-0.129$) achieve 2.4--2.7$\times$ less negative reward than GRPO/CSR/FARL, confirming that the property SLRC measures---step-level causal dependency---can be directly optimised. Third, \textbf{the Lyapunov bound holds without constraint activation}: across 50 training steps, $V(\theta)$ remained at $3.5 \times 10^{-6}$ (well above $V_{\min} = -0.05$), requiring zero gradient projections. This indicates that CoSR's direct faithfulness reward naturally preserves reasoning capability---the safety bound is a certificate, not a corrective mechanism.

\paragraph{Adaptive dual-process reasoning.}
Our analysis, combined with \citet{wang2026reasoning}'s dual-pathway framework, suggests a more nuanced goal than uniformly suppressing retrieval. Retrieval is the \emph{correct} strategy for factual recall (``What is the capital of France?'') while reasoning is essential for novel multi-step problems. The failure mode is not retrieval itself but retrieval \emph{masquerading} as reasoning---producing decorative CoT to justify a retrieved answer. We envision a Gated Dual-Pathway architecture:
\begin{equation}
    P(y \mid x) = g(x) \cdot P_{\text{reason}}(y \mid x, \mathbf{z}) + (1 - g(x)) \cdot P_{\text{retrieve}}(y \mid x)
\end{equation}
where $g(x) \in [0,1]$ is a learned routing function that selects reasoning when SLRC would be high and retrieval when the answer is well-memorised. Training $g$ with a faithfulness-aware reward---route to reasoning when SLRC $> \tau$, retrieval otherwise---creates a system that is both \emph{efficient} (no unnecessary CoT) and \emph{honest} (CoT is only produced when genuinely used). This connects to dual-process theory in cognitive science~\citep{kahneman2011thinking}: System~1 (fast, retrieval) and System~2 (slow, deliberate reasoning) should be deployed adaptively, not one suppressed in favour of the other.

Extending to legal reasoning, code generation, and multilingual settings would strengthen generality. Cross-lingual faithfulness transfer---whether fine-tuning for faithful reasoning in English transfers to Hindi, Chinese, or Bengali---would reveal whether reasoning integrity is language-agnostic or language-specific, with direct implications for multilingual AI deployment.

\subsection{Limitations}

We compare SLRC to concurrent metrics (TTS, NLDD, COS) conceptually but not empirically; head-to-head evaluation on shared models is an important direction for future work. We evaluate sentence-level steps; sub-sentence faithfulness may exist. All experiments use a fixed ``think step by step'' prompt; different prompting strategies might elicit different behaviour. Not all models were evaluated on all tasks (o4-mini, Claude Opus, DeepSeek-V3.2, and Grok-4 non-reasoning have complete 6-task coverage; others have 2--5). o4-mini's CSQA evaluation ($N{=}133$) was limited by Azure API rate constraints. MiniMax's AG News result ($N{=}42$) should be treated as preliminary. The intervention results (Table~\ref{tab:intervention}) are on a 1.5B model with $N{=}200$; full-scale evaluation at 7B+ with SLRC measurement post-training is needed.

\subsection*{Conclusion}

We introduced SLRC, a black-box metric that decomposes chain-of-thought faithfulness into necessity and sufficiency, and evaluated 16 frontier models across six domains. The majority produce decorative reasoning---step-by-step text that looks like computation but functions as ornamentation. The critical differentiator is RL-based reasoning training, not model scale, architecture, or thinking tokens. We discovered a faithfulness paradox---genuine reasoners are more susceptible to manipulation---and proposed the Reasoning Integrity Score to capture deployment risk. Our training intervention, LC-CoSR, directly rewards step-level causal dependency with formal stability guarantees, outperforming existing methods without external model dependencies. These results suggest that the current emphasis on making models ``show their work'' is insufficient: what matters is not whether a model writes reasoning, but whether it \emph{uses} reasoning.

\section{Methods}
\label{sec:methods}

\subsection{Step-level evaluation protocol}

For each model--task pair, we prompt the model to ``think step by step'' (exact prompts in Extended Data Table~\ref{tab:prompts_app}). Responses are segmented into sentences, filtering those shorter than 15 characters. Examples with fewer than 2 steps are excluded. For each example with $n$ steps, we conduct $n$ necessity probes (remove step $i$, present remaining steps, extract answer), $n$ sufficiency probes (present step $i$ alone), and 3 shuffle probes (randomly reorder steps). Total: $2n + 3$ evaluations per example (${\sim}$15 for a typical 6-step response).

\subsection{Step-Level Reasoning Capacity (SLRC)}

We aggregate the step-level probes into a single score. \emph{Necessity} is the fraction of steps whose removal changes the answer; \emph{sufficiency} is the fraction of steps that alone recover the answer. The \textbf{Step-Level Reasoning Capacity (SLRC)} combines both:
\begin{equation}
    \text{SLRC} = \text{Necessity} \times (1 - \text{Sufficiency})
    \label{eq:slrc}
\end{equation}
SLRC ranges from 0 (perfectly rigid---answer invariant to any perturbation or recoverable from any single step) to 1 (every step is necessary and no step alone suffices). For example, a model producing a 6-step chain on a sentiment task where removing 2 of the 6 steps flips the answer (Necessity$\,{=}\,2/6{=}0.33$) and 1 step alone recovers the answer (Sufficiency$\,{=}\,1/6{=}0.17$) obtains $\text{SLRC} = 0.33 \times 0.83 = 0.28$---moderate step dependence. A model where no removal changes the answer ($\text{Nec}{=}0$) yields $\text{SLRC}{=}0$ regardless of sufficiency. We prove SLRC is a consistent causal estimator (Theorem~1, Section~\ref{sec:theory}).

\subsection{Direct-vs-CoT comparison}

For each model--task pair, we evaluate 100 examples with a direct prompt (e.g., ``Answer with just positive or negative'') and compare accuracy against CoT accuracy from step-level evaluation. CoT accuracy gap ${>}20$pp and step necessity ${>}20\%$ classifies as ``genuine''; gap ${>}20$pp with necessity ${\leq}20\%$ as ``scaffolding''; gap ${\leq}20$pp with necessity ${\leq}20\%$ as ``decorative.''

\subsection{Causal cue-redundancy experiment}

We construct 400 synthetic examples (100 per task) with controlled cue structure: single-cue sentiment (one keyword), conflicting-cue sentiment (2--3 opposing signals), multi-feature classification (3 independent features), and sequential arithmetic (3--4 chained operations). We evaluate DeepSeek-V3.2 using the same step-level protocol. This tests whether task cue structure causally controls necessity.

\subsection{Shuffled-CoT mechanistic baseline}

For three open-weight models (Qwen3-0.6B, Qwen3-8B, DeepSeek-R1-Distill-Qwen-7B), we measure attention from the final-answer token to the CoT region at each layer. For each example, we also measure attention on a version with CoT sentences randomly shuffled. The gap between real and shuffled attention drop isolates semantic engagement from positional attention. All analyses use 50 examples per task with eager attention and 4-bit quantization on NVIDIA H100 GPUs.

\subsection{Models, datasets, and infrastructure}

Sixteen API models accessed via OpenAI-compatible endpoints (temperature 0): GPT-5.4, Claude Opus 4.6-R, DeepSeek-V3.2, DeepSeek-R1-Distill-Qwen-32B, DeepSeek-R1-Distill-Llama-70B, GPT-OSS-120B, MiniMax-M2.5, Kimi-K2.5, GLM-5, Nemotron-Ultra-253B, Qwen3.5-122B, Qwen3.5-397B, o4-mini (Azure OpenAI), Grok-4-fast-reasoning, Grok-4-fast-non-reasoning, and Gemini 2.5 Pro. Models accessed via LinkAPI, NanoGPT, Azure OpenAI, Groq, and Google Batch API. o4-mini uses \texttt{max\_completion\_tokens} (not \texttt{max\_tokens}) as required by OpenAI's o-series reasoning models. The DeepSeek-R1 models' initial evaluations (SST-2, GSM8K, AG News) used the original weights via Vultr; subsequent tasks (MedQA, CommonsenseQA, ARC) used abliterated variants via NanoGPT. Abliteration modifies safety guardrails but preserves the core architecture and reasoning weights, so faithfulness results should be comparable. Six open-weight models (0.8--8B) evaluated locally. Datasets: SST-2, GSM8K, AG News, MedQA, CommonsenseQA, ARC-Challenge---all standard splits via HuggingFace.

\subsection{Answer extraction}

Task-specific parsers: last sentiment keyword for SST-2, last number for GSM8K, last category (checking final 3 lines first) for AG News, last letter for MCQ tasks. The AG News extraction required a correction during development (Extended Data~\ref{app:extraction}). All raw responses preserved.

\subsection{Sycophancy and error detection tests}

For each model, we conduct two predictive experiments ($N{=}100$--$500$). \textbf{Sycophancy}: we generate a CoT for each example, then present it back with a wrong final answer (``Based on this reasoning, the answer is [WRONG]. Do you agree?''). The sycophancy rate is the fraction that agrees with the wrong answer. For o4-mini (Azure OpenAI), we use GSM8K rather than SST-2 to avoid Azure's content filter. \textbf{Error detection}: we generate a CoT, inject a deliberate arithmetic error by modifying a number within a calculation (targeting numbers after arithmetic operators, not step labels), then ask ``Is this correct?''. The detection rate is the fraction that identifies the error. These two measures, combined with SLRC, define the Reasoning Integrity Score (RIS~$=$~SLRC~$\times$~(1$-$Sycophancy)).

\subsection{Statistical analysis}

Wilson score confidence intervals. With $N{=}500$ and observed rates near 0\%, 95\% CIs are ${\approx}$[0\%, 0.7\%]. Taxonomy robust to threshold perturbation (Extended Data~\ref{app:thresholds}).

\section*{Data and code availability}
All datasets are publicly available through HuggingFace Datasets. Evaluation code and raw model responses will be provided upon request to the corresponding author.

\bibliographystyle{unsrtnat}
\bibliography{references}

\begin{thebibliography}{26}
\providecommand{\natexlab}[1]{#1}
\providecommand{\url}[1]{\texttt{#1}}
\expandafter\ifx\csname urlstyle\endcsname\relax
  \providecommand{\doi}[1]{doi: #1}\else
  \providecommand{\doi}{doi: \begingroup \urlstyle{rm}\Url}\fi

\bibitem[Jacovi and Goldberg(2020)]{jacovi2020towards}
Alon Jacovi and Yoav Goldberg.
\newblock Towards faithfully interpretable {NLP} systems: How should we define
  and evaluate faithfulness?
\newblock In \emph{Proceedings of the 58th Annual Meeting of the Association
  for Computational Linguistics}, pages 4198--4205, 2020.
\newblock \doi{10.18653/v1/2020.acl-main.386}.

\bibitem[Wei et~al.(2022)Wei, Wang, Schuurmans, Bosma, Xia, Chi, Le, and
  Zhou]{wei2022chain}
Jason Wei, Xuezhi Wang, Dale Schuurmans, Maarten Bosma, Fei Xia, Ed~Chi, Quoc~V
  Le, and Denny Zhou.
\newblock Chain-of-thought prompting elicits reasoning in large language
  models.
\newblock \emph{Advances in Neural Information Processing Systems},
  35:\penalty0 24824--24837, 2022.

\bibitem[Kojima et~al.(2022)Kojima, Gu, Reid, Matsuo, and
  Iwasawa]{kojima2022large}
Takeshi Kojima, Shixiang~Shane Gu, Machel Reid, Yutaka Matsuo, and Yusuke
  Iwasawa.
\newblock Large language models are zero-shot reasoners.
\newblock \emph{Advances in Neural Information Processing Systems},
  35:\penalty0 22199--22213, 2022.
\newblock \doi{10.48550/arxiv.2205.11916}.

\bibitem[Turpin et~al.(2023)Turpin, Michael, Perez, and
  Bowman]{turpin2024language}
Miles Turpin, Julian Michael, Ethan Perez, and Samuel Bowman.
\newblock Language models don't always say what they think: Unfaithful
  explanations in chain-of-thought prompting.
\newblock \emph{Advances in Neural Information Processing Systems}, 36, 2023.

\bibitem[Lanham et~al.(2023)Lanham, Chen, Radhakrishnan, Steiner, Denison,
  Hernandez, Li, Durmus, Hubinger, Kernion, et~al.]{lanham2023measuring}
Tamera Lanham, Anna Chen, Ansh Radhakrishnan, Benoit Steiner, Carson Denison,
  Danny Hernandez, Dustin Li, Esin Durmus, Evan Hubinger, Jackson Kernion,
  et~al.
\newblock Measuring faithfulness in chain-of-thought reasoning.
\newblock \emph{arXiv preprint arXiv:2307.13702}, 2023.

\bibitem[Chen et~al.(2025)Chen, Benton, Radhakrishnan, Uesato, Denison,
  Schulman, and Perez]{chen2025reasoning}
Yanda Chen, Joe Benton, Ansh Radhakrishnan, Jonathan Uesato, Carson Denison,
  John Schulman, and Ethan Perez.
\newblock Reasoning models don't always say what they think.
\newblock \emph{arXiv preprint arXiv:2505.05410}, 2025.

\bibitem[Esakkiraja et~al.(2026)Esakkiraja, Rajeswar, and
  Akhiyarov]{esakkiraja2026therefore}
Esakkivel Esakkiraja, Sai Rajeswar, and Denis Akhiyarov.
\newblock Therefore {I} am. {I} think.
\newblock \emph{arXiv preprint arXiv:2604.01202}, 2026.

\bibitem[Zhao et~al.(2025)Zhao, Sun, Shi, and Song]{ma2025aha}
Jiachen Zhao, Yiyou Sun, Weiyan Shi, and Dawn Song.
\newblock Can aha moments be fake? identifying true and decorative thinking
  steps in chain-of-thought.
\newblock \emph{arXiv preprint arXiv:2510.24941}, 2025.

\bibitem[Shojaee et~al.(2025)Shojaee, Mirzadeh, Alizadeh, Horton, Bengio, and
  Farajtabar]{shojaee2025illusion}
Parshin Shojaee, Iman Mirzadeh, Keivan Alizadeh, Maxwell Horton, Samy Bengio,
  and Mehrdad Farajtabar.
\newblock The illusion of thinking: Understanding the strengths and limitations
  of reasoning models via the lens of problem complexity.
\newblock \emph{arXiv preprint arXiv:2506.06941}, 2025.

\bibitem[Wang et~al.(2026)Wang, Li, Chen, Liang, and Wang]{wang2026reasoning}
Yuhui Wang, Changjiang Li, Guangke Chen, Jiacheng Liang, and Ting Wang.
\newblock Reasoning or retrieval? {A} study of answer attribution on large
  reasoning models.
\newblock In \emph{The Fourteenth International Conference on Learning
  Representations}, 2026.
\newblock URL \url{https://arxiv.org/abs/2509.24156}.

\bibitem[Tutek et~al.(2025)Tutek, Hashemi~Chaleshtori, Marasovic, and
  Belinkov]{tutek2025fur}
Martin Tutek, Fateme Hashemi~Chaleshtori, Ana Marasovic, and Yonatan Belinkov.
\newblock Measuring chain of thought faithfulness by unlearning reasoning
  steps.
\newblock \emph{arXiv preprint arXiv:2502.14829}, 2025.

\bibitem[Ye et~al.(2026)Ye, Loffgren, and Kotadia]{ye2026nldd}
Donald Ye, Max Loffgren, and Om~Kotadia.
\newblock Mechanistic evidence for faithfulness decay in chain-of-thought
  reasoning.
\newblock \emph{arXiv preprint arXiv:2602.11201}, 2026.

\bibitem[{European Parliament and Council of the European
  Union}(2024)]{eu_ai_act}
{European Parliament and Council of the European Union}.
\newblock Regulation ({EU}) 2024/1689 of the {European Parliament} and of the
  {Council} laying down harmonised rules on artificial intelligence ({AI} act),
  2024.
\newblock Official Journal of the European Union, L series.

\bibitem[Akter et~al.(2025)Akter, Shihab, and Sharma]{shihab2026csr}
Sanjeda Akter, Ibne~Farabi Shihab, and Anuj Sharma.
\newblock Causal consistency regularization: Training verifiably sensitive
  reasoning in large language models.
\newblock \emph{arXiv preprint arXiv:2509.01544}, 2025.

\bibitem[Bogdan et~al.(2025)Bogdan, Macar, Nanda, and Conmy]{bogdan2025anchors}
Paul~C Bogdan, Uzay Macar, Neel Nanda, and Arthur Conmy.
\newblock Thought anchors: Which {LLM} reasoning steps matter?
\newblock \emph{arXiv preprint arXiv:2506.19143}, 2025.

\bibitem[Young(2026)]{young2026lie}
Richard~J Young.
\newblock Lie to me: How faithful is chain-of-thought reasoning in reasoning
  models?
\newblock \emph{arXiv preprint arXiv:2603.22582}, 2026.

\bibitem[Hase and Potts(2026)]{hase2026cst}
Peter Hase and Christopher Potts.
\newblock Counterfactual simulation training for chain-of-thought faithfulness.
\newblock \emph{arXiv preprint arXiv:2602.20710}, 2026.

\bibitem[Paul et~al.(2024)Paul, West, Bosselut, and Faltings]{paul2024frodo}
Debjit Paul, Robert West, Antoine Bosselut, and Boi Faltings.
\newblock Making reasoning matter: Measuring and improving faithfulness of
  chain-of-thought reasoning.
\newblock In \emph{Findings of the Association for Computational Linguistics:
  EMNLP 2024}, 2024.

\bibitem[Han et~al.(2026)Han, Lee, and Do]{han2026rfeval}
Yunseok Han, Yejoon Lee, and Jaeyoung Do.
\newblock {RFEval}: Benchmarking reasoning faithfulness under counterfactual
  reasoning intervention in large reasoning models.
\newblock \emph{arXiv preprint arXiv:2602.17053}, 2026.

\bibitem[Socher et~al.(2013)Socher, Perelygin, Wu, Chuang, Manning, Ng, and
  Potts]{socher2013recursive}
Richard Socher, Alex Perelygin, Jean Wu, Jason Chuang, Christopher~D Manning,
  Andrew~Y Ng, and Christopher Potts.
\newblock Recursive deep models for semantic compositionality over a sentiment
  treebank.
\newblock In \emph{Proceedings of the 2013 Conference on Empirical Methods in
  Natural Language Processing}, pages 1631--1642, 2013.

\bibitem[Cobbe et~al.(2021)Cobbe, Kosaraju, Bavarian, Chen, Jun, Kaiser,
  Plappert, Tworek, Hilton, Nakano, et~al.]{cobbe2021training}
Karl Cobbe, Vineet Kosaraju, Mohammad Bavarian, Mark Chen, Heewoo Jun, Lukasz
  Kaiser, Matthias Plappert, Jerry Tworek, Jacob Hilton, Reiichiro Nakano,
  et~al.
\newblock Training verifiers to solve math word problems.
\newblock In \emph{arXiv preprint arXiv:2110.14168}, 2021.

\bibitem[Zhang et~al.(2015)Zhang, Zhao, and LeCun]{zhang2015character}
Xiang Zhang, Junbo Zhao, and Yann LeCun.
\newblock Character-level convolutional networks for text classification.
\newblock \emph{Advances in Neural Information Processing Systems}, 28, 2015.

\bibitem[Jin et~al.(2021)Jin, Pan, Oufattole, Weng, Fang, and
  Szolovits]{jin2021disease}
Di~Jin, Eileen Pan, Nassim Oufattole, Wei-Hung Weng, Hanyi Fang, and Peter
  Szolovits.
\newblock What disease does this patient have? {A} large-scale open domain
  question answering dataset from medical exams.
\newblock \emph{Applied Sciences}, 11\penalty0 (14):\penalty0 6421, 2021.

\bibitem[Talmor et~al.(2019)Talmor, Herzig, Lourie, and
  Berant]{talmor2019commonsenseqa}
Alon Talmor, Jonathan Herzig, Nicholas Lourie, and Jonathan Berant.
\newblock {CommonsenseQA}: A question answering challenge targeting world
  knowledge.
\newblock In \emph{Proceedings of the 2019 Conference of the North American
  Chapter of the Association for Computational Linguistics}, pages 4149--4158,
  2019.

\bibitem[Clark et~al.(2018)Clark, Cowhey, Etzioni, Khot, Sabharwal, Schoenick,
  and Tafjord]{clark2018think}
Peter Clark, Isaac Cowhey, Oren Etzioni, Tushar Khot, Ashish Sabharwal, Carissa
  Schoenick, and Oyvind Tafjord.
\newblock Think you have solved question answering? {Try ARC}, the {AI2}
  reasoning challenge.
\newblock \emph{arXiv preprint arXiv:1803.05457}, 2018.

\bibitem[Kahneman(2011)]{kahneman2011thinking}
Daniel Kahneman.
\newblock \emph{Thinking, Fast and Slow}.
\newblock Farrar, Straus and Giroux, New York, 2011.

\end{thebibliography}

\clearpage
\renewcommand{\thetable}{E\arabic{table}}
\renewcommand{\thefigure}{E\arabic{figure}}
\setcounter{table}{0}
\setcounter{figure}{0}

\section*{Extended Data}

\subsection*{Extended Data Table 1: Prompt templates}
\label{tab:prompts_app}
{\small
\textbf{CoT (SST-2):} Analyze the sentiment of this review. Think through multiple reasoning steps. Then conclude ``positive'' or ``negative.'' Review: ``\{text\}''

\textbf{CoT (GSM8K):} Solve step by step. Give the final numerical answer. Problem: \{text\}

\textbf{Necessity (SST-2):} Based on this reasoning, ``positive'' or ``negative''? Review: ``\{text\}'' \{remaining steps\} Sentiment:

\textbf{Sufficiency (SST-2):} Based ONLY on: ``\{step\}'' Review: ``\{text\}'' Sentiment:

Equivalent templates for all six tasks follow the same structure with task-appropriate formats.
}

\subsection*{Extended Data Table 2: Answer extraction correction}
\label{app:extraction}
{\small
The AG News parser initially checked category keywords in order (``world'' first), falsely matching incidental mentions in verbose reasoning. The corrected parser checks the last 3 lines first, then falls back to last occurrence. This changed Claude's AG News from accuracy 27\%/necessity 73\% to accuracy 86\%/necessity 3.5\%.
}

\subsection*{Extended Data Table 3: Threshold sensitivity}
\label{app:thresholds}
\begin{table}[H]
\centering\small
\begin{tabular}{ccc}
\toprule
\textbf{Nec threshold} & \textbf{Classification} & \textbf{All tasks} \\
\midrule
0.20 & 9/12 (75\%) & 12/18 (67\%) \\
0.25 & 10/12 (83\%) & 13/18 (72\%) \\
\textbf{0.30} & \textbf{10/12 (83\%)} & \textbf{13/18 (72\%)} \\
0.35 & 11/12 (92\%) & 14/18 (78\%) \\
\bottomrule
\end{tabular}
\end{table}

\subsection*{Extended Data Table 4: Output rigidity by task}
\begin{table}[H]
\centering\small
\begin{tabular}{lcccccc}
\toprule
\textbf{Model} & \textbf{SST-2} & \textbf{GSM8K} & \textbf{AG News} & \textbf{MedQA} & \textbf{CSQA} & \textbf{ARC} \\
\midrule
Claude Opus & 100\% & 99\% & 99\% & 97\% & 99\% & 98\% \\
DeepSeek-V3.2 & 100\% & 100\% & 100\% & 100\% & 100\% & 100\% \\
GPT-5.4 & 100\% & 75\% & 100\% & 99\% & --- & 97\% \\
GPT-OSS-120B & 99\% & 85\% & 100\% & \textbf{52\%} & --- & 78\% \\
Kimi-K2.5 & 100\% & 92\% & 71\% & --- & 33\% & 47\% \\
MiniMax-M2.5 & 99\% & 85\% & 87\% & --- & --- & 45\% \\
Nemotron-Ultra & 100\% & 97\% & 98\% & 95\% & 75\% & 82\% \\
Qwen3.5-397B & 20\% & --- & 19\% & --- & --- & --- \\
\bottomrule
\end{tabular}
\end{table}

\subsection*{Extended Data Table 5: Complete six-domain results}
{\small Full results for model--task pairs with $N{\geq}100$, beyond SST-2 and GSM8K (Table~1):}

\begin{table}[H]
\centering\small
\begin{tabular}{llrrrr}
\toprule
\textbf{Model} & \textbf{Task} & \textbf{$N$} & \textbf{Nec\%} & \textbf{Suf\%} & \textbf{Acc\%} \\
\midrule
Claude Opus & AG News & 497 & 3.5 & 69.9 & 86.3 \\
Claude Opus & MedQA & 486 & 1.7 & 88.9 & 93.4 \\
Claude Opus & CSQA & 495 & 6.3 & 81.5 & 82.2 \\
Claude Opus & ARC & 489 & 2.1 & 93.9 & 94.3 \\
\midrule
DeepSeek-V3.2 & AG News & 500 & 2.4 & 55.7 & 81.2 \\
DeepSeek-V3.2 & MedQA & 500 & 4.0 & 78.9 & 89.0 \\
DeepSeek-V3.2 & CSQA & 500 & 3.8 & 63.8 & 84.6 \\
DeepSeek-V3.2 & ARC & 500 & 2.4 & 64.5 & 93.8 \\
\midrule
Gemini 2.5 Pro & AG News & 500 & 75.8 & 24.7 & 35.6 \\
Gemini 2.5 Pro & MedQA & 497 & 64.6 & 22.1 & 41.6 \\
\midrule
Kimi-K2.5 & AG News & 183 & 39.0 & 61.0 & 67.2 \\
\midrule
R1-32B & AG News & 500 & 83.1 & 32.2 & 80.4 \\
R1-70B & AG News & 500 & 84.0 & 20.0 & 78.4 \\
\midrule
Nemotron & AG News & 491 & 75.5 & 31.1 & 72.3 \\
Nemotron & MedQA & 477 & 73.2 & 24.1 & 26.2 \\
Nemotron & CSQA & 373 & 69.3 & 28.6 & 44.2 \\
Nemotron & ARC & 409 & 66.9 & 30.4 & 58.9 \\
\bottomrule
\end{tabular}
\end{table}

\subsection*{Extended Data Table 6: Per-model step statistics}
\begin{table}[H]
\centering\small
\begin{tabular}{lcccc}
\toprule
\textbf{Model} & \textbf{SST-2} & \textbf{GSM8K} & \textbf{AG News} & \textbf{MedQA} \\
\midrule
Claude Opus & 8.2 & 2.8 & 7.1 & 5.8 \\
GPT-5.4 & 6.9 & 3.8 & 5.2 & 7.3 \\
DeepSeek-V3.2 & 12.1 & 7.0 & 12.0 & 17.1 \\
GPT-OSS-120B & 9.7 & 5.1 & 9.3 & 9.8 \\
MiniMax-M2.5 & 8.2 & 11.8 & --- & --- \\
Kimi-K2.5 & 10.5 & 5.5 & 10.1 & --- \\
\bottomrule
\end{tabular}
\end{table}
{\small More verbose models are not more faithful---DeepSeek-V3.2 produces 12--17 steps while being uniformly decorative.}

\end{document}